# Causality Extraction from Nuclear Licensee Event Reports Using a Hybrid Framework

Shahidur Rahoman Sohag [†], Sai Zhang*, Min Xian[†], Shoukun Sun[†], Fei Xu*, Zhegang Ma*

[†]Department of Computer Science, University of Idaho, 1776 Science Center Dr., Idaho Falls, ID 83406, mxian@uidaho.edu
*Idaho National Laboratory, 1955 Fremont Ave., Idaho Falls, ID 83405, Sai.Zhang@inl.gov

## INTRODUCTION

Industry-wide nuclear power plant operating experience is a critical source of raw data for performing parameter estimations in reliability and risk models. Much operating experience information pertains to failure events and is stored as reports containing unstructured data, such as narratives. Event reports are essential for understanding how failures are initiated and propagated, including the numerous causal relations involved.

Causal relation extraction using deep learning represents a significant frontier in the field of natural language processing (NLP), and is crucial since it enables the interpretation of intricate narratives and connections contained within vast amounts of written information. Such capabilities are vital in various applications, ranging from decision making and predictive analytics to enhancing our understanding of natural language. Its advanced representation learning capabilities make deep learning the technique best suited for detecting and comprehending subtle patterns and hidden connections in text. It allows text to be analyzed with a degree of intricacy and nuance closely resembling human comprehension, thereby facilitating the development of NLP applications that are increasingly precise and perceptive. However, implementing deep learning for cause-and-effect relationship extraction is fraught with challenges. One major challenge is the subtlety with which these relationships are often expressed in text, requiring models to discern intricate patterns and contextual nuances. In addition, deep learning models require substantial computational resources and large datasets with annotated examples for training, and these can be resource-intensive to compile and process.

Causal relations can be categorized into three groups [1]: (1) explicit causality, which refers to a direct causal relationship in which the cause and effect are clearly stated and connected using language patterns; (2) implicit causality, which occurs when the cause-and-effect relationship can only be inferred from the surrounding context or the verb choices; and (3) embedded causality, which occurs when the causal relation is not part of the main clause in a sentence but is included within a subordinate clause, providing additional context or information. Various methodologies have been suggested for identifying causes and effects, including knowledge-based, statistical machine-learning-based, and deep-learning-based approaches [2-4]. Our study (Fig. 1) involves integrating deep learning with knowledge-based approaches via the use of hybrid methods. Our methodology consists of two distinct stages: causality classification and cause-effect extraction. The first stage classifies sentences into two categories (binary classification), based on whether they express cause-and-effect relationships in textual data. This approach employs a combination of NLP techniques and deep learning architectures in order to analyze and interpret the data. Researchers from Idaho National Laboratory prepared a dataset with 343 cause-effect pairs from an U.S. nuclear power plant operating experience database [6]. The second stage focuses on extracting specific cause-effect segments from the sentences via a sophisticated keyword-centric methodology.

## METHODS

### Data and Materials

The raw data for this study included 92 licensee event reports (LERs) submitted to the U.S. Nuclear Regulatory Commission (NRC) by domestic nuclear power plant licensees, and that had been made publicly available on the NRC website. For LERs, the proposed approach selects the following 12 fields for processing: facility name, title, event date, report date, cause, system, component, manufacturer, reportability to the Industry Reporting and Information System, abstract, event description, and cause description. Note that only the information in three of the fields (i.e., abstract, event description, and cause description) is used to learn causal relations. The information in the other fields is extracted as a placeholder for potential future studies (e.g., report classification and report-to-report relevancy). The selected 92 LERs all pertained to motor-driven pump failures.

### Data Preprocessing

Two nuclear experts annotated the cause-effect pairs from raw LER documents by using a web-based interactive tool developed in this work. A total of 343 cause-effect pairs were collected from the 92 LERs. The text cleaning process removes and normalizes specific unwanted characters and whitespaces. First, it detects newline characters ('\n' and '\\n') and replaces them with a space that eliminates unwanted sentence breaks due to formatting issues. Additionally,

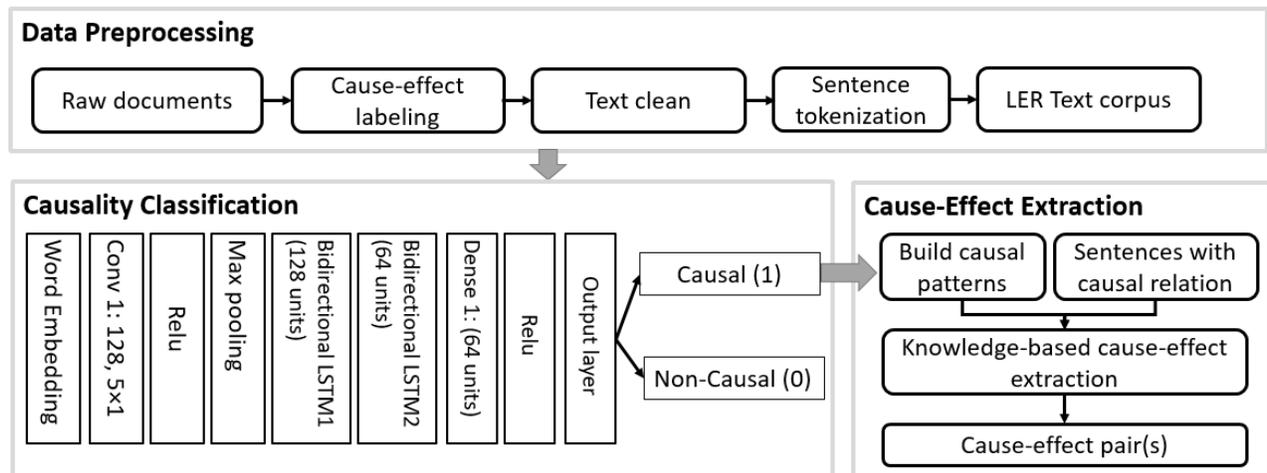

Fig. 1. Pipeline of the proposed method. Conv: convolutional layer. ReLU: rectified linear units

Unicode replace characters ('\ufffd') and Unicode escape sequences ('\\u[0-9a-fA-F]{4}') are replaced with spaces. This is essential for preserving text data integrity, particularly when working with multiple data sources that may have different encoding standards. Furthermore, all whitespaces with a single space are removed, along with any tabs with many consecutive spaces. Finally, the leading and trailing whitespaces from the sentences are trimmed during sentence splitting and tokenization.

The sentence tokenization step generates individual sentences from documents. The LER corpus contains structural text data prepared for model development (e.g., machine learning training and evaluation). A record in the LER corpus is defined as $(S_i, l_i, C_i, E_i)$, where $S_i$ is a piece of text containing up to three consecutive sentences, $l_i \in \{'causal', 'non-causal'\}$ denotes the type/label of $S_i$, and $C_i$ and $E_i$ represent the specific cause event and effect event described in $S_i$, respectively. If $l_i$ is $'non-causal'$, both $C_i$ and $E_i$ are set to 'NA'. Ultimately, the LER corpus consisted of 20,129 records.

**Causality Classification**

The proposed deep learning model begins with an embedding layer that transforms the integer-encoded sentences into dense vectors. The embedding layer converts these numbers into a format that captures the context of each word within an output dimension of 150, while managing a large vocabulary. Following the embedding layer is a convolutional layer [7] with 128 filters and a kernel size of 5. It is activated by a ReLU function [8]. This layer scans five-word sequences for distinctive textual features or patterns. Batch normalization [9] is then performed for more stable training, followed by a MaxPooling layer to reduce the dimensionality of the data. The heart of the proposed model lies in the two bidirectional long short-term memory layers [10]. These are adept at remembering and using both past (backward) and future (forward) information in a sentence. The first layer has 128 units and passes on its full analysis; the second, with 64 units, condenses the information further. Dropout layers are used to prevent overfitting [11], and dense layers with ReLU activation further process the extracted features. The final layer is a dense layer with a single neuron and a sigmoid activation function, suitable for the binary classification task at hand. The model is compiled with the Adam optimizer [12] and binary cross-entropy loss function, and is trained on the padded training data, with early stopping implemented to prevent overfitting. This comprehensive architecture ensures the model not only learns to identify patterns indicative of causal relations, but also performs well when applied to novel, unfamiliar data.

In the LER text corpus, text samples could contain as many as three sentences, allowing the model to capture longer causal contexts that may span multiple sentences.

**Cause-Effect Extraction**

The second stage focuses on extracting precise cause-effect pairs from the sections of text containing causal relations. This is accomplished via a refined pattern-based approach. Two groups of causal patterns were defined, as shown in TABLE I. The C-E type defines patterns with cause and effect on the left and right segments of the sentence, respectively. The E-C type defines them in the opposite manner. First, we inputted the sections of text containing causal relations. The algorithm searched for the location of any causal patterns within the text, then determined the cause and effect segments according to the pattern type. Once these segments were identified, the algorithm paired them to form distinct cause-effect relations. This two-stage approach not only detects the existence of causality but also defines the precise sections of text that reflect such relationships. By integrating the pattern-based technique with the deep learning model, this methodology serves as a reliable tool for analyzing intricate textual data and generating significant cause-effect insights.

TABLE I. Causal patterns

| Pattern type | Patterns | No. |
|---|---|---|
| **E-C** | "as a result of," "attributable to," "based on," "because of," "cause of," "caused by," "determined," "determined by," "determined that," "determined to be," "discovered that," "due to," "failure mechanism," "found," "given that," "identified," "indications," "initiated by," "on account of," "owing to," "rendering," "resulted from," "revealed," "source," "stemmed from," "triggered by" | 26 |
| **C-E** | "attributed to," "caused," "causes," "causing," "generated," "indicated," "indicates that," "lead to," "leading to," "leads to," "resulted in," "the reason for," "yielded" | 13 |

TABLE II. Causality classification results. 0 denotes non-causal class and 1 denotes causal class.

| | | Samples | Precision | Recall | F1 |
|---|---|---|---|---|---|
| Training set | 0 | 15,868 | 1.00 | 0.99 | 1.0 |
| | 1 | 234 | 0.81 | 1.0 | 0.90 |
| | 0+1 | 16,102 | Avg. accuracy: 0.997 | | |
| Test set | 0 | 3,981 | 1.00 | 1.0 | 1.0 |
| | 1 | 45 | 0.77 | 0.91 | 0.84 |
| | 0+1 | 4,026 | Avg. accuracy: 0.991 | | |

## RESULTS

### Causality Classification

The proposed deep learning model was specifically developed to detect any sentence(s) containing causal relations. Eighty percent of the samples from the LER text corpus were used as a training set, while the remaining 20% were used for testing. Since the causal and non-causal classes are extremely imbalanced, resampling was performed to ensure an equal number of samples for both classes. Four metrics—precision, recall ratio, F1-score, and average accuracy—were used to evaluate model performance.

In TABLE I, the causality classification model achieved high average accuracies of 99.7% and 99.1% on the training and test sets, respectively. The recall ratios for causal sentence detection on both the training and test sets were also high (100% and 91.0%). The lower precision values (81.0% and 77.0%) indicate that a relatively high number of non-causal samples were misclassified as causal samples. The difference resulted from our design goal of uncovering all possible causal relations, even at the cost of higher false positives.

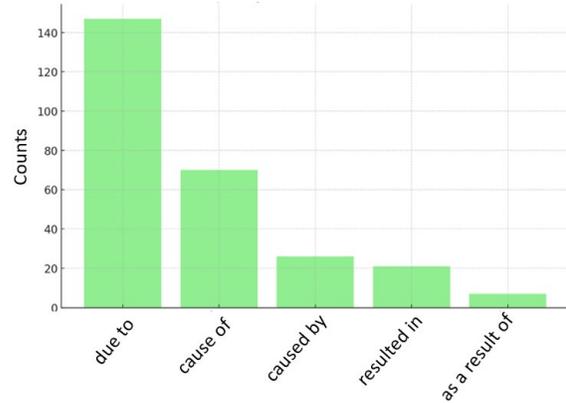

Fig. 2. Occurrences of major causal patterns in the LER corpus.

The proposed model obtained high precision values (100%) for non-causal samples on both the training and test sets, demonstrating its potential for effectively excluding text containing non-causal relations. Furthermore, the findings shown in TABLE II highlight the model's ability to effectively manage the imbalanced data.

### Cause-Effect Extraction

In this section, 252 of the total 279 causal samples were included to validate the proposed approach. The other 27 causal samples were excluded because they contained no explicitly or embedded causality.

Fig. 2 shows the occurrence of causal patterns in the LER corpus. "Due to" is the most frequently used pattern that indicates a causal relation, with "cause of" coming in a rather distant second. "Caused by," "resulted in," and "as a result of" round out the list. We handcrafted 39 causal patterns (see TABLE I) in order to extract causal relations from the text corpus. After identifying these phrases, the algorithm extracted text segments indicative of cause and effect, organizing them into pairs based on their contextual relationships.

Sample results obtained using the knowledge-based extraction strategy (see Table III) indicate a high degree of accuracy for single-sentence scenarios. The algorithm accurately identified and matched the expected causes and accompanying effects. For example, "inertial latch binding" is accurately identified as the cause of the "DB-50 supply breaker to Auxiliary Feedwater Pump 21 not closing," and "the foam ring had deteriorated" is the cause behind "a piece of foam being drawn into the suction piping of the 'A' MDAFW pump." The extracted cause-effect pairs were manually compared against the ground truth in the corpus, showing that the proposed approach correctly identified 181 cause-effect pairs out of the total 252. The main errors that arose were from pairing causes and effects in embedded causal sentences. Considering that the dataset was small, with a limited number of manually labeled cause-effect pairs, the performance of the proposed approach was reasonable. The results emphasize that future effort is needed for better extracting and matching cause-effect pairs in terms of implicit and embedded causality.

TABLE III. Examples of cause-effect extraction. MDAFW: motor-driven auxiliary feedwater. ECW: essential cooling water. MDEFW: motor-driven emergency feedwater.

| Input sample | Cause-effect pair (cause, effect) |
| --- | --- |
| The DB-50 supply breaker to Auxiliary Feedwater Pump 21 did not close due to inertial latch binding | (inertial latch binding, the db-50 supply breaker to auxiliary feedwater pump 21 did not close) |
| The foam ring had deteriorated causing a piece of the foam to tear loose and be drawn into the suction piping of the "A" MDAFW pump. | (The foam ring had deteriorated, a piece of the foam to tear loose and be drawn into the suction piping of the "A" MDAFW pump) |
| The bearing degradation was due to insufficient tolerance in the motor shaft endplay, as set during refurbishment. Corrective actions are planned to specify this design parameter for subsequent refurbishments, and to increase endplay adjustment shim thickness in the affected ECW pump motors to reduce bearing wear. | (insufficient tolerance in the motor shaft endplay, as set during refurbishment, the bearing degradation) |
| A subsequent cause evaluation determined that a similar 2B MDEFW pump trip on February 1, 2021 was also caused by the intermittent poor electrical connection due to the loose shorting screws. | (The intermittent poor electrical connection, The loose shorting screws) (A similar 2B MDEFW pump trip on February 1, 2021, The intermittent poor electrical connection) |

**CONCLUSION**

This paper proposed a hybrid framework for causality detection and extraction from LER documents. The main contributions include: (1) we compiled an LER corpus with 20,129 text samples for causality analysis, (2) developed an interactive tool for labeling cause-effect pairs, (3) built a deep-learning-based approach for causal relation detection, and (4) developed a knowledge-based cause-effect extraction approach. In the future, we will increase the text corpus size, label more causal samples, and develop and evaluate large language models usable for causality extraction.